\documentclass[runningheads]{llncs}

\usepackage{accv}

\usepackage{accvabbrv}

\usepackage{graphicx}
\usepackage{booktabs}
\usepackage{bm}
\usepackage{multirow}
\usepackage{rotating}
\usepackage{tabularx}
\usepackage{placeins}
\usepackage{subcaption}

\usepackage[accsupp]{axessibility}  

\usepackage{hyperref}

\usepackage{orcidlink}

\begin{document}

\title{Federated Class Incremental Learning: A Pseudo Feature Based Approach Without Exemplars} 

\titlerunning{ FCLPF}

\author{Min Kyoon Yoo\inst{1}\orcidlink{0009-0007-2149-3341} \and
Yu Rang Park\inst{1}\orcidlink{0000-0002-4210-2094} }

\authorrunning{M. K. Yoo and Y. R. Park}

\institute{Department of Biomedical Systems Informatics, Yonsei University College of Medicine, Seoul, Republic of Korea\\
\email{minkyoon\_da@yonsei.ac.kr, yurangpark@yuhs.ac}\\}

\maketitle

\begin{abstract}
  Federated learning often assumes that data is fixed in advance, which is unrealistic in many real-world scenarios where new data continuously arrives, causing catastrophic forgetting. To address this challenge, we propose FCLPF (Federated Class Incremental Learning with Pseudo Features), a method that uses pseudo features generated from prototypes to mitigate catastrophic forgetting. Our approach reduces communication costs and improves efficiency by eliminating the need for past data and avoiding computationally heavy models like GANs. Experimental results on CIFAR-100 show that FCLPF achieves an average accuracy of 51.87\% and an average forgetting of 9.62\%, significantly outperforming existing baselines with an average accuracy of 47.72\% and forgetting of 20.46\%. On TinyImageNet, FCLPF achieves 37.56\% accuracy and 3.14\% forgetting, compared to the baselines' 27.69\% accuracy and 24.46\% forgetting, demonstrating the superior performance of FCLPF. 
\end{abstract}

\section{Introduction}
\label{sec:intro}

Federated learning \cite{konevcny2016federated} is decentralized machine learning that does not require data centralization . The absence of centralized data collection allows for the preservation of privacy. It has recently attracted considerable attention in both research and industry. It is employed in fields such as healthcare \cite{qayyum2022healthcarefl} and the development of local devices \cite{imteaj2021survey}.

In conventional federated learning, it is assumed that data remains fixed over time. This assumption is not valid in the real world. For example, in healthcare, A disease classification model may require incorporating new disease data over time. Therefore, the model must be capable of learning new data without forgetting previously learned information.

Catastrophic forgetting \cite{mccloskey1989catastrophic} occurs when the addition of new data causes the model to forget previously learned information due to class imbalance. In the centralized setting, this issue is studied under incremental learning \cite{schlimmer1986incremental}. Several methods have been proposed to mitigate catastrophic forgetting \cite{castro2018end, hou2019learning, rebuffi2017icarl,zhao2020maintaining}.

Federated class incremental learning (FCIL) \cite{dong2022federated} addresses the aforementioned challenges by preventing catastrophic forgetting when learning new data in a federated learning environment. A significant proportion of the existing proposed methods make use of rehearsal memory \cite{dong2022federated,dong2023no,psaltis2023fedrcil}. Nevertheless, past data retention can prove challenging for devices with limited storage \cite{ravaglia2021tinyml}, potentially raising concerns about data privacy.

To address these issues, we propose FCLPF (Federated Class Incremental Learning with Pseudo Features), a novel approach that generates pseudo features without relying on rehearsal memory. The proposed FCLPF generates pseudo features by utilizing only prototypes and data in the client. To create a pseudo feature, only prototype types need to be exchanged between the client and server, reducing communication costs and preserving privacy. This method prevents catastrophic forgetting and is efficient in federated learning environments. The code is available on GitHub\footnote{\url{https://github.com/DigitalHealthcareLab/24FCLPF.git}}.

\section{Related Work}
\noindent\textbf{Federated Class Incremental Learning.} The objectives of Federated Class Incremental Learning (FCIL) is to preserve previously learned knowledge while incorporating new data into the global model.

The studies GLFC \cite{dong2022federated} and LGA \cite{dong2023no} employ a proxy server to mitigate catastrophic forgetting. However, having a proxy server raises possible privacy issues and increases communication costs. Furthermore, contrary to common federated learning assumptions, both studies allow clients to share data. Moreover, rehearsal memory is necessary for both methods, which is impractical for devices with little storage.   Additionally,  FedRCIL \cite{psaltis2023fedrcil} employs self-supervised learning (SSL) and knowledged distillation to FCIL, yet it also leverages rehearsal memory.

 In certain instances, past data is not accessible, and the retention of such data is challenging for devices with limited memory. Several methods of using GAN \cite{goodfellow2014generative} to replace past data have been proposed. FedCIL \cite{qi2023better} generates synthetic data for global model consolidation by using locally trained generative models. MFCL \cite{babakniya2024data} and TARGET \cite{zhang2023target} also demonstrate the feasibility of FCIL without the use of rehearsal memory. However, MFCL and TARGET employ a centralized server to train GAN, which incurs a significant computational cost and a high communication cost for sending and receiving models.

\bigskip 

\noindent\textbf{Exemplar-Free Class Incremental Learning}
Class incremental learning (CIL) algorithms can be employed effectively when new data is introduced. Catastrophic forgetting occurs when learning new data and is a significant challenge in machine learning. This is due to the lack of past data, a crucial factor in the learning process. Several CIL methods have recently been developed to combat this problem \cite{castro2018end, hou2019learning, rebuffi2017icarl,zhao2020maintaining}. Numerous exemplar-based approaches have been proposed. However, these methods present challenges in low-memory environments and may raise privacy concerns.

To address these challenges, Exemplar-Free CIL (EFCIL) has been proposed as a potential solution \cite{yu2020semantic, zhu2021prototype, smithalways}. These methods do not retain past data. A significant number of EFCIL methods employ regularization techniques to facilitate incremental updates to the deep model, and distillation is used to preserve past knowledge \cite{masana2022class}.

Some techniques, such as PASS \cite{zhu2021prototype}, employ prototype augmentation to prevent catastrophic forgetting across incremental states; other techniques, such as ABD \cite{smithalways}, use picture inversion to produce pseudo-samples of previous classes. It then uses linear head fine-tuning and importance-weighted feature distillation to distinguish between task features from the past and present without the requirement for rehearsal memory.

For IL2A \cite{zhu2021class}, the features are generated using the information from the class distribution to create the features of previous classes. In FeTrIL \cite{petit2023fetril}, the feature extractor is fixed, and pseudo-features are generated straightforwardly. Our method fixes the feature extractor like FeTrIL and generates pseudo features. By fixing the feature extractor, only the fully connected (FC) layer needs to be exchanged, which significantly reduces the communication cost, making this an efficient and suitable method in federated class incremental learning. To the best of our knowledge, our study is the first to apply pseudo features to FCIL.

\section{Method}

\begin{figure}
    \centering
    \includegraphics[width=0.7\linewidth]{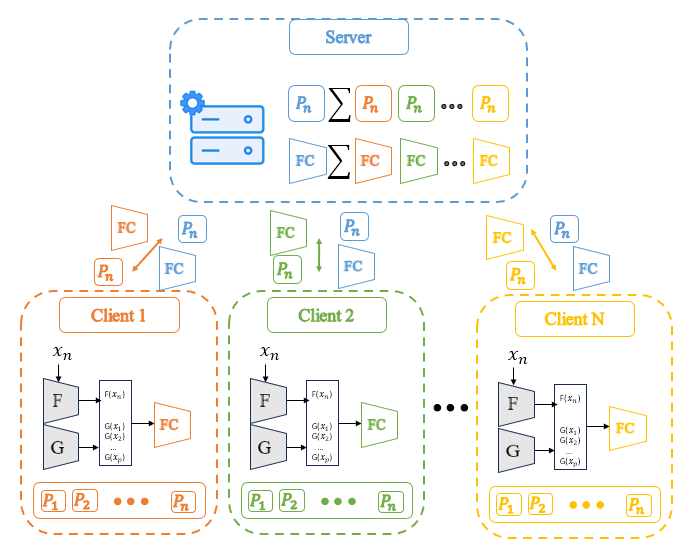}
    \caption{Overview of our method. 
    (\( P_n \)) is the prototype of the \(n\)th class, and FC denotes the fully connected (FC) layer. \(F(x_n)\) represents the feature of \(x_n\), whereas \(G(x_p)\) denotes the pseudo feature for the past class. Only prototypes and FC layers are exchanged between the client and server. The server aggregates the prototype and FC layer.
    }
    \label{fig:enter-label}
\end{figure}

We propose FCLPF, a federated class incremental learning that uses pseudo features to replace rehearsal memory. Previous studies are mostly with rehearsal memory, and those without rehearsal memory create and use GANs, which leads to very high computational and communication costs. Therefore, we created pseudo features to replace rehearsal memory. This study is the first to apply pseudo features to FCIL. We propose a new way to create pseudo features using Principal Component Analysis (PCA) \cite{wold1987principal}. Our method is illustrated in Figure \ref{fig:enter-label}.

Initially, the client trains the feature extractor and the FC layer. At the end of the initial task, the server aggregates the class-specific prototypes from all clients, performing a sample-weighted prototype averaging based on the number of samples per class for each client. This ensures that the class distribution of all clients is reflected in the global model.

Following the initial task, the feature extractor is frozen. The server sends only the FC layers and global prototypes to the clients. The clients use the received prototypes to generate pseudo features, replacing rehearsal memory. This approach reduces the parameters that need to be shared with the central server by sharing only FC layers and prototypes, significantly reducing communication costs in a federated learning environment.

\bigskip 

\noindent\textbf{Prototype.} We don't save the past data. We create a pseudo feature. To create the pseudo feature, we used the prototype. A prototype was created using the following formula:

\begin{equation}
\bm{\mu}_c = \frac{1}{\left| X_c \right|} \sum_{\bm{x} \in X_c} f_c(\bm{x})
\end{equation}

\noindent ,where \( f_c(\bm{x}) \)  denotes feature vector of sample x belonging to class \( c \). \( \bm{\mu}_c \) denotes prototype for class \( c \). We extracted the features for each class and averaged them. We sent the pseudo features created by each client in the first round of the task to the center and averaged them to create a single prototype. The prototype created by the center was sent to each client so that they could create pseudo features within the client.

\bigskip

\noindent\textbf{Closest prototype.} We compute the new class that is most similar to the old class we want to create. The closest prototype was calculated using the formula as follows:

\begin{equation}
C_{\text{closest}} = \arg\min_{c_{\text{new}} \in C_{\text{new}}} \text{dist}(\bm{\mu}_{c_{\text{old}}}, \bm{\mu}_{c_{\text{new}}})
\end{equation}

\noindent, where \(C_{\text{new}}\) represents the set of new classes, and \(\bm{\mu}_{c_{\text{old}}}\) and \(\bm{\mu}_{c_{\text{new}}}\) represents the prototype of the old class and the prototype of the new class, respectively. \(\text{dist}(\cdot, \cdot)\) denotes the distance metric. In this study, we employed the cosine similarity as the distance metric. We then calculated the difference between the old prototype \( \bm{\mu}_{c_{\text{old}}}\) and the closest new class prototype \( \bm{\mu}_{c_{\text{closest}}}\):

\begin{equation}
\Delta_{c_{\text{old}}, c_{\text{closest}}} = \bm{\mu}_{c_{\text{old}}} - \bm{\mu}_{c_{\text{closest}}}.
\end{equation}

\noindent\textbf{PCA Projection.} We applied PCA to the prototype of the old class \(\bm{\mu}_{c_{\text{old}}}\), the prototypes of all new classes \(\bm{\mu}_{C_{\text{new}}}\) to find the principal axes. We projected \(\Delta_{c_{\text{old}}, c_{\text{closest}}}\) onto these principal axes and restored the original dimensions as follows:

\begin{equation}
\text{PCA}_k(\{\bm{\mu}_{c_{\text{old}}}\} \cup \{\bm{\mu}_{C_{\text{new}}}\}) = P_k
\end{equation}

\begin{equation}
\Delta_{c_{\text{old}}, c_{\text{closest}}}^{\text{projected}} = \left(\Delta_{c_{\text{old}}, c_{\text{closest}}} \times P_k^T\right) \times P_k
\end{equation}

\noindent, where \( P_k \) represents the matrix of the \( k \) principal components that explains the most variance.
\bigskip

\noindent\textbf{Pseudo feature generation.} We generated an old class pseudo feature. We added the \(\Delta_{c_{\text{old}}, c_{\text{closest}}}^{\text{projected}}\) to \(f_{\text{closest}}(\bm{x})\). We created a pseudo feature as follows:

\begin{equation}
f_{c_{\text{old}}}(\bm{x}) = f_{C_{\text{closest}}}(\bm{x}) + \Delta_{c_{\text{old}}, c_{\text{closest}}}^{\text{projected}}
\end{equation}

\noindent, where \(f_{c_{\text{old}}}(\bm{x})\) represents the pseudo feature of the old class generated by using \(f_{\text{closest}}(\bm{x})\) which is the feature vector of sample \(\bm{x}\) belonging to class \(C_{\text{closest}}\).

\section{Experiment}

\noindent\textbf{Setting.}  We experimented with our method CIFAR-100 \cite{krizhevsky2009learning}, TinyImageNet \cite{le2015tiny}, and SuperImageNet \cite{babakniya2024data} to demonstrate its effectiveness. We used Resnet18 \cite{he2016deep} as the backbone model and compared our model with Fedavg \cite{mcmahan2017communication}, FedProx \cite{li2020federated}, and  MFCL \cite{babakniya2024data}. The parameter settings are identical to those described in MFCL \cite{babakniya2024data}. For CIFAR-100, the number of clients is 50, and the local epoch is 10. For TinyImageNet and SuperImageNet, the number of clients is 100, and the local epoch is 10. The SGD optimizer was utilized. The data for each task was distributed among the clients using Latent Dirichlet Allocation (LDA) \cite{reddi2020adaptive} with $\alpha = 1$. The learning rate for all tasks starts at 0.1 and decreases exponentially to 0.01 over time.   
\bigskip

\noindent\textbf{Incremental setting.} We configured the experiment the same as the previous Exemplar-free class incremental learning setting \cite {zhu2021class, zhu2021prototype, zhu2022self}. The CIFAR-100 dataset was tested in three scenarios:

\begin{description}
    \item[\textbf{Scenario 1:}] \makebox[0.9\linewidth][l]{50 initial classes and 5 incremental learning states of 10 classes each}
    \item[\textbf{Scenario 2:}] \makebox[0.9\linewidth][l]{50 initial classes and 10 incremental learning states of 5 classes each}
    \item[\textbf{Scenario 3:}] \makebox[0.9\linewidth][l]{40 initial classes and 20 incremental learning states of 3 classes each}
\end{description}

\noindent The TinyImageNet dataset was also tested in three scenarios:

\begin{description}
    \item[\textbf{Scenario 1:}] \makebox[0.9\linewidth][l]{100 initial classes and 5 incremental learning states of 20 classes each}
    \item[\textbf{Scenario 2:}] \makebox[0.9\linewidth][l]{100 initial classes and 10 incremental learning states of 10 classes each}
    \item[\textbf{Scenario 3:}] \makebox[0.9\linewidth][l]{100 initial classes and 20 incremental learning states of 5 classes each}
\end{description}

\noindent The SuperImageNet dataset was additionally tested in two separate scenarios:

\begin{description} \item[\textbf{Scenario 1:}] \makebox[0.9\linewidth][l]{30 initial classes and 5 incremental learning states of 4 classes each} \item[\textbf{Scenario 2:}] \makebox[0.9\linewidth][l]{30 initial classes and 10 incremental learning states of 2 classes each} \end{description}

\bigskip

\noindent\textbf{Metric.} We compared the performance metric with average accuracy. Average accuracy averages the accuracy of all tasks. It is calculated as follows:

\begin{equation}
\widetilde{A} = \frac{1}{T} \sum_{t=1}^{T} A_t
\end{equation}
where \( A_t \) is the accuracy of the task \( t \), \( T \) is the total number of tasks, and \( \widetilde{A} \) is the average accuracy.

Average forgetting evaluates how much the model forgets the previous task while learning a new task. Forgetting (\(f_t\)) of task \(t\) is defined as the difference between the highest performance on task \(t\) and the performance after all training. The average forgetting is calculated by averaging all the 
\(f_t\) for tasks \(1\) to \(T-1\) Using the following formula:
\begin{equation}
\widetilde{f} = \frac{1}{T-1} \sum_{t=1}^{T-1} f_t
\end{equation}

\subsection{Results}

\begin{figure*}[tb]
    \centering
    \begin{subfigure}[b]{0.4\textwidth}
        \includegraphics[width=\textwidth]{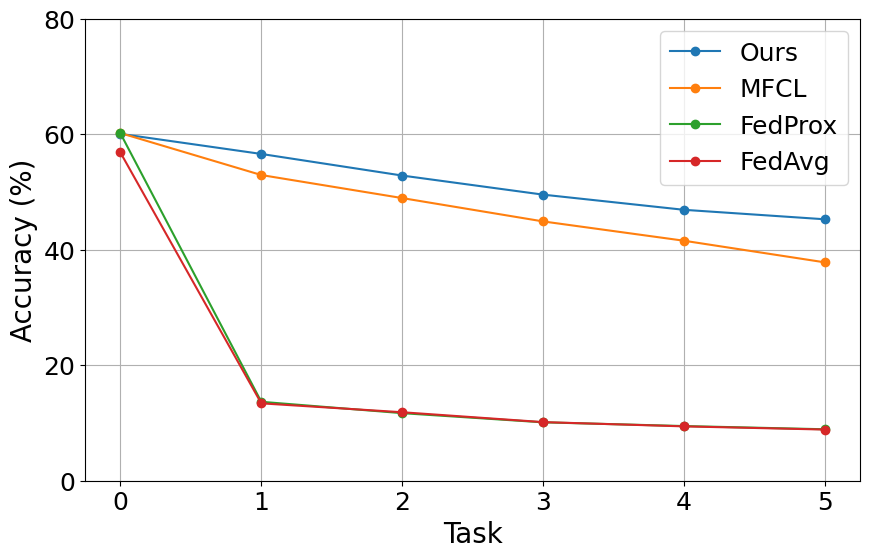}
        \caption{CIFAR Scenario 1}
        \label{fig:CIFAR_T5}
    \end{subfigure}
    \begin{subfigure}[b]{0.4\textwidth}
        \includegraphics[width=\textwidth]{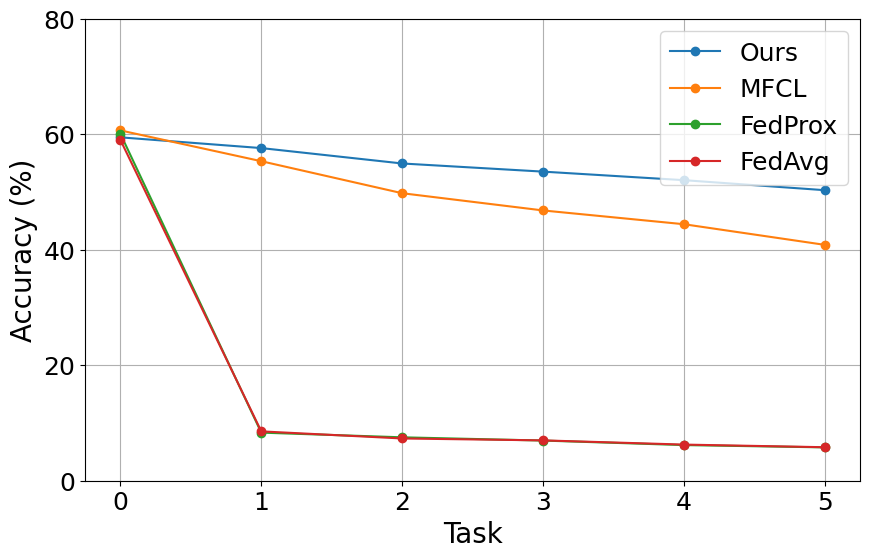}
        \caption{CIFAR Scenario 2}
        \label{fig:CIFAR_T10}
    \end{subfigure}
    \begin{subfigure}[b]{0.4\textwidth}
        \includegraphics[width=\textwidth]{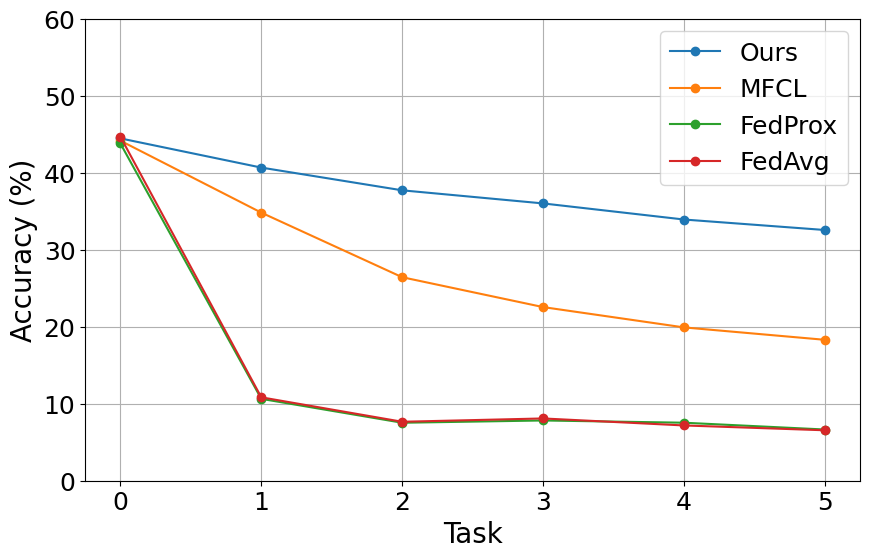}
        \caption{TinyImageNet Scenario 1}
        \label{fig:TinyImageNet_T5}
    \end{subfigure}
    \begin{subfigure}[b]{0.4\textwidth}
        \includegraphics[width=\textwidth]{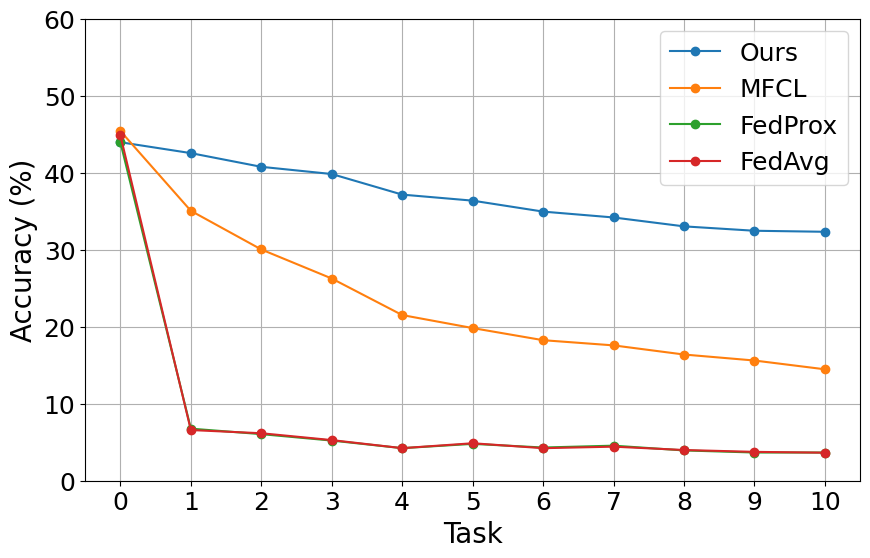}
        \caption{TinyImageNet Scenario 2}
        \label{fig:TinyImageNet_T10}
    \end{subfigure}
    \caption{Comparison of model performance across different tasks and datasets}
    \label{fig:comparison}
\end{figure*}

The proposed method demonstrated superior performance compared to the baseline both datasets across all scenarios. Figure \ref{fig:comparison} illustrates the performance for each task. The method demonstrated the highest performance for all tasks. The detailed experimental results are presented in Table \ref{tab:comparison_cifar}, \ref{tab:comparison_tinyimagenet}, and \ref{tab:comparison_superimagenet} , which compare the average accuracy and average forgetting across different scenarios. Our method outperformed the other methods in all scenarios.

In the CIFAR-100 dataset, the average accuracy of FCLPF is 51.87\%, 51.06\%, and 49.30\% for scenarios 1, 2, and 3, respectively. In contrast, the other federated class incremental learning method, MFCL, achieved 47.72\%, 43.17\%, and 37.13\%, respectively. The performance gap between our method and the other methods widened as the number of tasks increased. Our method achieved the best average forgetting results. FCLPF's average forgetting was significantly lower, with values of 3.14\%, 4.00\%, and 4.10\% in scenarios 1, 2, and 3, respectively. This indicates better retention of previous knowledge compared to MFCL, which had average forgetting values of 20.46\%, 33.57\%, and 43.73\% in the same scenarios.

\begin{table*}[tb]
  \caption{Comparison of Average Accuracy and Average Forgetting for CIFAR-100}
  \label{tab:comparison_cifar}
  \centering
  \footnotesize
  \renewcommand{\arraystretch}{0.7}
  \setlength{\tabcolsep}{12pt} 
  \begin{tabular}{@{}l l c c@{}}
    \toprule
    Scenario & Method & Average Accuracy (\%) & Average Forgetting (\%) \\
    \midrule
    \multirow{4}{*}{S1} 
    & \textbf{Ours}     & \textbf{51.87 ± 5.23} & \textbf{9.62 ± 2.47} \\
    & MFCL     & 47.72 ± 7.40 & 20.46 ± 29.63 \\
    & FedAvg   & 18.42 ± 17.25 & 65.97 ± 32.74 \\
    & FedProx  & 19.00 ± 18.47 & 66.12 ± 32.79 \\
    \midrule
    \multirow{4}{*}{S2} 
    & \textbf{Ours}     & \textbf{51.06 ± 4.65} & \textbf{10.35 ± 4.52} \\
    & MFCL     & 43.17 ± 8.78 & 33.57 ± 28.02 \\
    & FedAvg   & 10.79 ± 15.29 & 79.74 ± 26.49 \\
    & FedProx  & 10.90 ± 15.59 & 80.04 ± 26.52 \\
    \midrule
    \multirow{4}{*}{S3} 
    & \textbf{Ours}     & \textbf{49.30 ± 5.40} & \textbf{10.88 ± 6.95} \\
    & MFCL     & 37.13 ± 9.57 & 43.73 ± 22.18 \\
    & FedAvg   & 6.73 ± 12.02 & 85.99 ± 20.24 \\
    & FedProx  & 6.74 ± 12.06 & 85.71 ± 20.49 \\
    \bottomrule
  \end{tabular}
\end{table*}

In Tiny ImageNet, our proposed FCLPF demonstrated superior performance compared to other baselines. Furthermore, the performance difference between our method and other baselines increased with the increased number of tasks. In Tiny ImageNet, the average accuracy of our method was 37.56\%, 37.05\%, and 37.01\% in scenarios 1, 2, and 3, respectively. In contrast, MFCL achieved 27.69\%, 23.67\%, and 16.55\%, respectively. The average forgetting is also dominant over MFCL, and as the task size increases, our method's average forgetting becomes more pronounced compared to other baselines.

In SuperImageNet, our proposed FCLPF showed superior performance compared to other baselines. In Scenario 1, our method achieved an average accuracy of 65.37\%, while MFCL only achieved 51.21\%. Similarly, in Scenario 2, our method recorded 65.10\%, significantly outperforming MFCL's 47.44\%. Moreover, our method consistently exhibited lower average forgetting compared to other baselines. In Scenario 1, our method had an average forgetting of only 8.36\%, while MFCL's forgetting reached 28.41\%. Similarly, in Scenario 2, our method showed an average of 11.67\% forgetting, while MFCL's forgetting was much higher at 40.87\%. These results confirm that FCLPF is more effective in mitigating forgetting and maintaining higher accuracy in SuperImageNet scenarios than other baselines.

\begin{table*}[tb]
  \caption{Comparison of Average Accuracy and Average Forgetting for TinyImageNet}
  \label{tab:comparison_tinyimagenet}
  \centering
  \footnotesize
  \renewcommand{\arraystretch}{0.7}
  \setlength{\tabcolsep}{12pt} 
  \begin{tabular}{@{}l l c c@{}}
    \toprule
    Scenario & Method & Average Accuracy (\%) & Average Forgetting (\%) \\
    \midrule
    \multirow{4}{*}{S1} 
    & \textbf{Ours}     & \textbf{37.56 ± 4.04} & \textbf{3.14 ± 2.89} \\
    & MFCL     & 27.69 ± 9.12 & 24.46 ± 21.33 \\
    & FedAvg   & 14.16 ± 13.71 & 49.67 ± 24.99 \\
    & FedProx  & 14.00 ± 13.39 & 49.47 ± 25.01 \\
    \midrule
    \multirow{4}{*}{S2} 
    & \textbf{Ours}     & \textbf{37.05 ± 3.95} & \textbf{4.00 ± 3.47} \\
    & MFCL     & 23.67 ± 9.24 & 38.06 ± 20.61 \\
    & FedAvg   & 8.36 ± 11.58 & 63.30 ± 21.39 \\
    & FedProx  & 8.28 ± 11.32 & 63.24 ± 21.41 \\
    \midrule
    \multirow{4}{*}{S3} 
    & \textbf{Ours}     & \textbf{37.01 ± 3.68} & \textbf{4.10 ± 4.17} \\
    & MFCL     & 16.55 ± 8.00 & 48.18 ± 18.88 \\
    & FedAvg   & 4.67 ± 9.02 & 73.64 ± 17.69 \\
    & FedProx  & 4.61 ± 8.85 & 73.46 ± 17.85 \\
    \bottomrule
  \end{tabular}
\end{table*}

\begin{table*}[tb]
  \caption{Comparison of Average Accuracy and Average Forgetting for SuperImageNet}
  \label{tab:comparison_superimagenet}
  \centering
  \footnotesize
  \renewcommand{\arraystretch}{0.7}
  \setlength{\tabcolsep}{12pt} 
  \begin{tabular}{@{}l l c c@{}}
    \toprule
    Scenario & Method & Average Accuracy (\%) & Average Forgetting (\%) \\
    \midrule
    \multirow{4}{*}{S1} 
    & \textbf{Ours}     & \textbf{65.37 ± 5.37} & \textbf{8.36 ± 3.33} \\
    & MFCL     & 51.21 ± 11.04 & 28.41 ± 33.99 \\
    & FedAvg   & 18.99 ± 23.87 & 67.36 ± 34.93 \\
    & FedProx  & 19.04 ± 23.74 & 68.43 ± 35.42 \\
    \midrule
    \multirow{4}{*}{S2} 
    & \textbf{Ours}     & \textbf{65.10 ± 5.43} & \textbf{11.67 ± 5.32} \\
    & MFCL     & 47.44 ± 10.92 & 40.87 ± 33.34 \\
    & FedAvg   & 8.36 ± 11.58 & 63.30 ± 21.39 \\
    & FedProx  & 10.63 ± 19.44 & 78.64 ± 29.38 \\
    \bottomrule
  \end{tabular}
\end{table*}

\begin{figure*}[tb]
    \centering
    \includegraphics[width=0.8\textwidth]{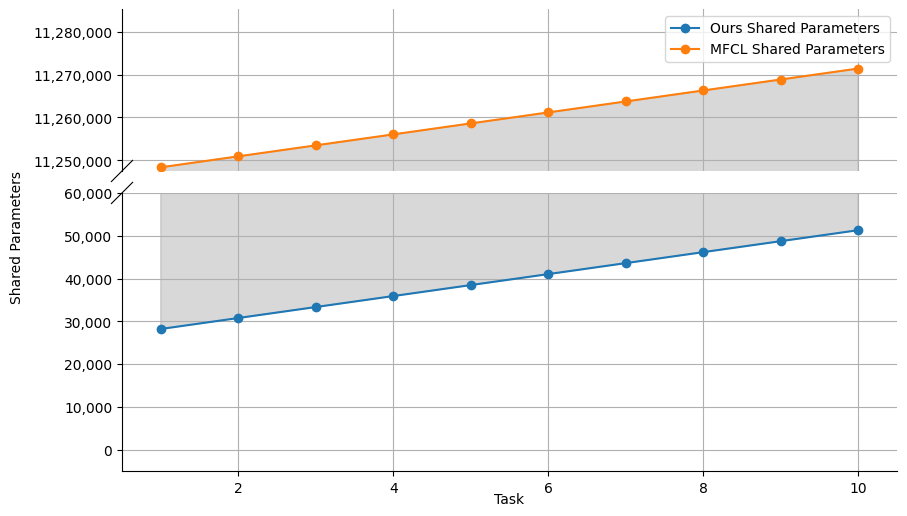}
    \caption{Comparison of shared parameters across tasks for different methods. The figure illustrates the number of shared parameters per round for each task.}
    \label{fig:shared_parameters}
\end{figure*}

Figure \ref{fig:shared_parameters} compares the typical number of parameters exchanged per round for each task by method. As the initial task parameters are similar, we compare the shared parameters after the initial task. In contrast to other methods, our approach does not exchange feature extractors, only FC layer. Furthermore, since we employ pseudo features, we are not required to exchange models, such as GANs, which is a significant advantage. Consequently, our method is more efficient in terms of communication costs.

The experimental results demonstrate that FCLPF achieves high accuracy and is an effective method to prevent catastrophic forgetting. Furthermore, the minimal number of shared parameters makes it a highly efficient algorithm.
\FloatBarrier

 \section{Conclusion}
In this study, we proposed FCLPF (Federated Class Incremental Learning With Pseudo Features), a novel method that employs pseudo features in federated class incremental learning for privacy-preserving and communication efficiency without relying on rehearsal memory. Our approach leverages PCA to generate pseudo features. Our experimental results on CIFAR-100 and TinyImageNet showed that our model is superior in all tasks compared to other baseline models. The key innovation of FCLPF is creating pseudo features to replace rehearsal memory, resulting in low communication costs since only prototypes need to be exchanged. The results presented in this study demonstrate that the proposed method prevents catastrophic forgetting and facilitates efficient federated learning.
 
For future work, we plan to explore training the feature extractor in conjunction with pseudo features to enhance the model's performance. Furthermore, even though we presently average the prototypes, our goal is to explore different approaches for creating and applying prototypes to improve the accuracy and robustness of the pseudo features. 

Overall, FCLPF represents a significant advancement in federated class incremental learning. The study addresses the limitations of rehearsal memory and presents an efficient method for reducing communication costs while using pseudo-features.

\bigskip
 
\noindent\textbf{Acknowledgement.}This research was supported by the Korea Institute for Advancement of Technology (KIAT) grant funded by the Korea Government (MOTIE) (No. P0023675, HRD Program for Industrial Innovation) and a grant of the Korea Machine Learning Ledger Orchestration for Drug Discovery Project (K-MELLODDY), funded by the Ministry of Health \& Welfare and Ministry of Science and ICT, Republic of Korea (Grant Number: RS-2024-00459866).

%
%
\bibliographystyle{splncs04}

\bibliography{main}
\end{document}